\documentclass[sigconf]{acmart}
\usepackage{balance}
\AtBeginDocument{%
  }

\copyrightyear{2025}
\acmYear{2025}
\setcopyright{acmlicensed}
\acmConference[MM '25] {Proceedings of the 33rd ACM International Conference on Multimedia}{October 27--31, 2025}{Dublin, Ireland.}
\acmBooktitle{Proceedings of the 33rd ACM International Conference on Multimedia (MM '25), October 27--31, 2025, Dublin, Ireland}
\acmISBN{979-8-4007-2035-2/2025/10}
\acmDOI{10.1145/3746027.3755474}

\newcommand{\ourmethod}{AnchorSync}
\begin{document}

%%
%% The "title" command has an optional parameter,
%% allowing the author to define a "short title" to be used in page headers.

\title{\ourmethod: Global Consistency Optimization for Long Video Editing}

%%
%% The "author" command and its associated commands are used to define
%% the authors and their affiliations.
%% Of note is the shared affiliation of the first two authors, and the
%% "authornote" and "authornotemark" commands
%% used to denote shared contribution to the research.
\author{Zichi Liu}
%\authornote{Both authors contributed equally to this research.}
\email{liuzichi@sjtu.edu.cn}
\orcid{0009-0008-7154-4220}
% \author{G.K.M. Tobin}
% \authornotemark[1]
% \email{webmaster@marysville-ohio.com}
\affiliation{%
  \institution{MoE Key Lab of Artificial Intelligence, AI
Institute\\ Shanghai Jiao Tong University}
  \city{Shanghai}
  \country{China}
}

\author{Yinggui Wang}
\email{wyinggui@gmail.com}
\orcid{0000-0002-6686-6603}
\affiliation{%
  \institution{Ant Group}
  \city{Shanghai}
  \country{China}}

\author{Tao Wei}
\email{lenx.wei@antgroup.com}
\orcid{0009-0000-4027-0310}
\affiliation{%
  \institution{Ant Group}
  \city{Shanghai}
  \country{China}}

\author{Chao Ma}
\email{chaoma@sjtu.edu.cn}
\orcid{0000-0002-8459-2845}
\authornote{~Corresponding author.}
\affiliation{%
  \institution{MoE Key Lab of Artificial Intelligence, AI
Institute\\ Shanghai Jiao Tong University}
  \city{Shanghai}
  \country{China}}

%%
%% By default, the full list of authors will be used in the page
%% headers. Often, this list is too long, and will overlap
%% other information printed in the page headers. This command allows
%% the author to define a more concise list
%% of authors' names for this purpose.
%\renewcommand{\shortauthors}{Trovato et al.}

%%
%% The abstract is a short summary of the work to be presented in the
%% article.
\begin{abstract}
  Editing long videos remains a challenging task due to the need for maintaining both global consistency and temporal coherence across thousands of frames. Existing methods often suffer from structural drift or temporal artifacts, particularly in minute-long sequences. We introduce \ourmethod, a novel diffusion-based framework that enables high-quality, long-term video editing by decoupling the task into sparse anchor frame editing and smooth intermediate frame interpolation. Our approach enforces structural consistency through a progressive denoising process and preserves temporal dynamics via multimodal guidance. Extensive experiments show that AnchorSync produces coherent, high-fidelity edits, surpassing prior methods in visual quality and temporal stability. The source code is available at \href{https://github.com/VISION-SJTU/AnchorSync}{https://github.com/VISION-SJTU/AnchorSync}.
\end{abstract}

%%
%% The code below is generated by the tool at http://dl.acm.org/ccs.cfm.
%% Please copy and paste the code instead of the example below.
%%


%%
%% Keywords. The author(s) should pick words that accurately describe
%% the work being presented. Separate the keywords with commas.
%\vspace{-0.2cm}
%% A "teaser" image appears between the author and affiliation
%% information and the body of the document, and typically spans the
%% page.
\begin{teaserfigure}
  \centering
  \includegraphics[width=.98\textwidth]{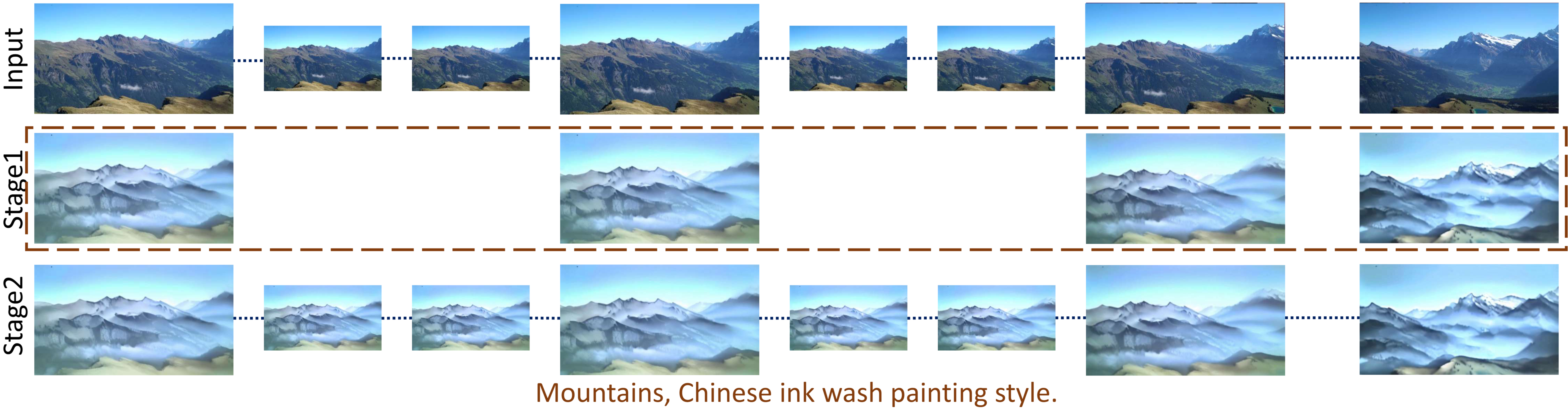}
  \vspace{-4mm}
  \caption{\textbf{Overview of our two-stage framework.}  
We first perform consistent editing on sparsely sampled anchor frames in Stage 1.  
Then, we smoothly generate the intermediate frames in Stage 2. 
% via multimodal-guided interpolation (Stage 2), ensuring temporal smoothness and structural continuity across the full sequence.
}
  % \Description{Enjoying the baseball game from the third-base
  % seats. Ichiro Suzuki preparing to bat.}
  \label{fig:teaser}
  \vspace{2mm}
\end{teaserfigure}

% \received{14 April 2025}
% \received[revised]{19 June 2025}
% \received[accepted]{4 July 2025}

%%
%% This command processes the author and affiliation and title
%% information and builds the first part of the formatted document.
%\renewcommand\footnotetextcopyrightpermission[1]{}
%\settopmatter{printacmref=false} %remove ACM reference format
\maketitle

% \vspace{-0.3cm}
\section{Introduction}

Video editing is a fundamental task in visual content creation, enabling applications from film post-production and virtual reality to data augmentation in several machine learning tasks ~\cite{detection-1,detection-2,segmentation-1,segmentation-2,tracking-1,tracking-2}. With the rise of diffusion models, image editing has seen tremendous progress in both quality and controllability~\cite{null-text-inversion,imagefeature-pnp, imagefeature-p2p,imagefeature-zero,style-1,style-2,style-3,t2iapplication-blended}. However, extending them to video editing presents unique challenges due to the need to maintain both spatial fidelity and temporal consistency across frames.

Most existing diffusion-based video editing methods are designed for short clips, typically under 100 frames~\cite{i2vedit,i2vedit-revideo,tokenflow,oneshot-tuneavideo,oneshot-towards,zeroshot-videop2p,zeroshot-text2video,zeroshot-vid2vid}. This limitation arises from high GPU memory requirements and the increasing difficulty of ensuring coherent structure and motion over longer time spans. 

Two dominant strategies have emerged to address these challenges: frame-by-frame editing and segment-by-segment editing. The former processes each frame independently while maintaining temporal features across time~\cite{t2vedit-streamv2v}, which struggles with long-term consistency. The latter divides videos into shorter chunks and performs localized editing. Depending on how segments are connected, their performance varies. Some methods~\cite{i2vedit-flowvid,i2vedit-anyv2v,i2vedit-ccedit} condition each segment on the result of the previous one to maintain local consistency, but this design inevitably introduces cumulative errors across segments.
Rerender~\cite{zeroshot-rerender}, on the other hand, edits the anchor frames individually without joint optimization, making it susceptible to inconsistencies across keyframes. It then employs Ebsynth for frame propagation, which is prone to failure under complex motion, leading to structural artifacts and temporal flickering.
More recently, Gen-L-Video~\cite{gen-l-video} applies inversion-based editing on overlapping video segments and blends overlapping regions at each denoising step. While this approach reduces boundary artifacts, it relies on short-video editing models with limited temporal awareness, often resulting in inconsistent appearances and unsmooth transitions in long video sequences.

In this paper, we propose \textbf{\ourmethod}, a novel diffusion-based framework for long video editing that explicitly tackles long-term consistency and short-term continuity in a unified architecture. Our approach decouples the editing process into two stages: (1) \textit{anchor frame editing}, where a sparse set of representative frames are jointly edited through a progressive denoising process. To ensure global coherence, we inject a trainable Bidirectional Attention into a diffusion model to capture pairwise structural dependencies between distant frames, and perform Plug-and-Play (PnP) inversion and injection for controllable editing; and (2) \textit{intermediate frame interpolation}, where we leverage a video diffusion model equipped with a newly trained multimodal ControlNet to guide generation using both optical flow and edge maps, enabling temporally smooth and structure-aware transitions between anchor frames.
Extensive experiments on videos up to several minutes long demonstrate that our method produces edits with significantly higher temporal consistency, better visual fidelity, and smoother transitions compared to state-of-the-art methods. %Our framework is scalable, flexible, and achieves robust performance across diverse scenes, motions, and editing prompts.
Our main contributions are summarized as follows:
\begin{itemize}
\setlength\itemsep{0mm}
    \item We propose \textbf{\ourmethod}, a unified two-stage framework for long video editing that jointly addresses long-term consistency and short-term continuity.
    \item To ensure global consistency, we introduce a bidirectional attention mechanism for enforcing pairwise frame consistency, along with a progressive pairwise diffusion strategy to propagate coherence across all anchor frames.
    \item For interpolation, we train a multimodal ControlNet for SVD with Bidirectional Temporal Frame Fusion to efficiently generate smooth, coherent transitions.
\end{itemize}

\begin{figure*}[tp]
    \centering
    \includegraphics[width=.92\linewidth]{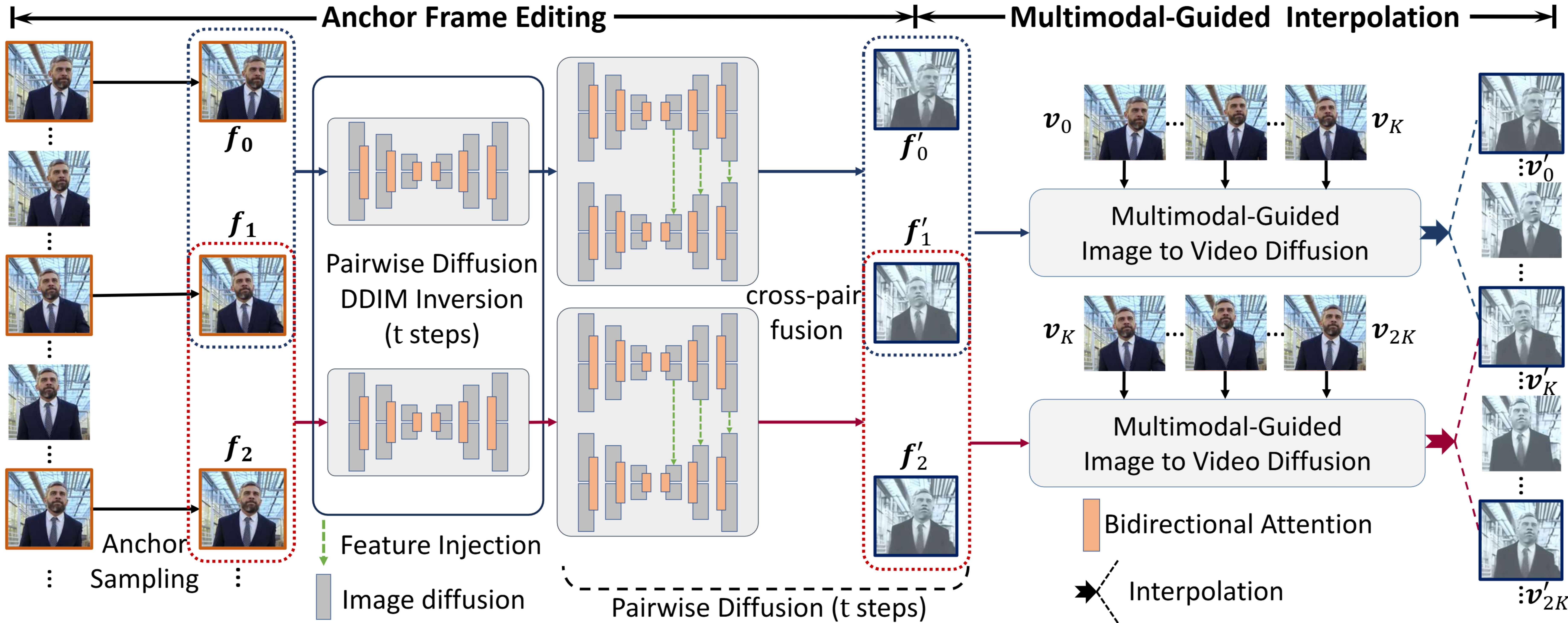}
    \vspace{-3mm}
    \caption{Overview of \ourmethod{}. %(A) \textbf{Video Editing Pipeline}. 
    The framework consists of two stages: \textbf{(1) Anchor Frame Editing}, where sparse keyframes are first inverted via DDIM inversion (pairwise diffusion) and then jointly edited using pairwise diffusion. Each anchor frame is edited with its adjacent frames using \textit{Bidirectional Attention} and feature injection, ensuring structure-aware consistency across time. \textbf{(2) Multimodal-Guided Interpolation}, where intermediate frames are generated by a multimodal-guided image-to-video diffusion model, conditioned on both optical flow and edge cues. This design enables high-quality editing for long videos while maintaining global consistency and temporal smoothness.
    }
    \label{fig:method}
    \vspace{-2mm}
\end{figure*}

\section{Related Work}

{\flushleft\bf Diffusion-based Image Editing.}
Diffusion models~\cite{DM-1, DM-2, t2i-diffusion2, t2i-glide, t2i-hierarchical, t2i-scene} have achieved significant breakthroughs in image editing tasks, enabling high-quality synthesis through controllable denoising processes. Early approaches, such as InstructPix2Pix~\cite{instructpix2pix}, enabled semantic-level edits via instruction-conditioned fine-tuning. ControlNet~\cite{controlnet} further introduced spatial guidance (e.g., edges, depth maps) into the diffusion pipeline, enhancing structural control during generation.
To improve structural fidelity without fine-tuning the entire model, Plug-and-Play (PnP) methods~\cite{imagefeature-pnp} perform latent inversion and inject editing guidance during denoising. These methods have proven effective for structure-preserving image transformation. In our work, we extend PnP-style editing to anchor frames in long videos. Additionally, we adopt LoRA~\cite{lora} as an efficient parameter-tuning strategy during the training of mutual attention modules.

% Recently, Plug-and-Play (PnP) methods~\cite{tumanyan2023pnp, wang2023composable} have emerged as a flexible paradigm for image editing. These methods perform inversion to reconstruct a latent representation of the input image and then apply iterative guidance injections during denoising to steer the output toward a desired edit. PnP techniques are particularly attractive for preserving structure while achieving semantic transformation, without requiring model fine-tuning. Our method builds upon this line of work, integrating a PnP-style strategy into a video editing pipeline to improve global consistency and structure retention.
{\flushleft\bf Diffusion-based Video Editing.}
While diffusion models have shown promising results in video generation and editing, most existing methods ~\cite{zeroshot-videop2p,zeroshot-pix2video,zeroshot-rerender,oneshot-towards,oneshot-tuneavideo} are limited to short clips, typically no longer than 100 frames. Editing long videos remains challenging due to substantial GPU memory consumption, temporal drift over time, and difficulties in maintaining structural and motion consistency. As a result, only a few recent works have attempted to scale diffusion-based editing to minute-long sequences.
To manage these challenges, current methods mainly follow two paradigms: frame-by-frame editing and segment-by-segment editing. Frame-by-frame methods, such as StreamV2V~\cite{t2vedit-streamv2v}, sequentially edit each frame using temporal feature fusion and caching but often suffer from flickering and lack of global semantic consistency. 
Segment-based approaches divide the video into short chunks for localized editing. Rerender~\cite{zeroshot-rerender} edits only the first frame of each segment and propagates the effect using Ebsynth, which leads to structural inconsistency under large motion. Anyv2v~\cite{i2vedit-anyv2v} performs conditional editing based on the first frame; we adapt it for long videos by conditioning each segment on the last frame of the previous one, but this causes cumulative errors across segments. Gen-L-Video~\cite{gen-l-video} extends short-video editing models by applying inversion-based editing to overlapping segments, using weighted averaging for transition regions. However, due to the limited temporal awareness of short-video models, the results still exhibit inconsistency and unsmooth transitions in long videos.

To overcome these limitations, we propose a unified two-stage framework that edits anchor frames via a consistency-aware denoising process and interpolates intermediate frames using a multimodal-guided diffusion model. This enables scalable, high-quality editing of videos with thousands of frames.

\vspace{-1mm}
\section{Method}

\subsection{Overview}
Given a video sequence $\mathcal{V} = \{\boldsymbol{v}_i\}_{i=0}^{T}$ with $T$ frames, our goal is to generate an edited version $\mathcal{V}' = \{\boldsymbol{v}'_i\}_{i=0}^{T}$ that reflects the editing prompt $\boldsymbol{y}^{edit}$, while preserving the spatial structure and temporal dynamics of the original video. To facilitate Plug-and-Play (PnP) editing in the latent space of diffusion models, we also introduce an inversion prompt $\boldsymbol{y}^{inv}$, which provides semantic grounding for reconstructing the original content prior to editing.

% 位置不够下面可以删
Editing long videos introduces significant challenges: direct frame-by-frame editing often leads to flickering and drift, while existing short-video editing methods struggle to scale to thousands of frames with consistent semantics. 

To address these issues, we propose a two-stage framework that decouples long video editing into two manageable yet coherent phases. (1) we sample a sparse set of anchor frames from $\mathcal{V}$ and apply a joint editing process based on pairwise diffusion. Each anchor frame is edited in coordination with its neighbors using a Bidirectional Attention mechanism, ensuring structure-consistent modifications aligned with the prompt $\boldsymbol{y}^{edit}$. (2) we interpolate the intermediate frames between edited anchors using a multimodal-guided image-to-video diffusion model. By conditioning on low-level structural cues such as Canny edges and optical flow, the model generates smooth and temporally consistent transitions that preserve both motion dynamics and edited semantics.

Together, this two-stage design enables scalable, high-fidelity editing for minute-long videos while maintaining global coherence throughout the sequence.

\begin{figure}[tp]
    \centering
    \includegraphics[width=.9\linewidth]{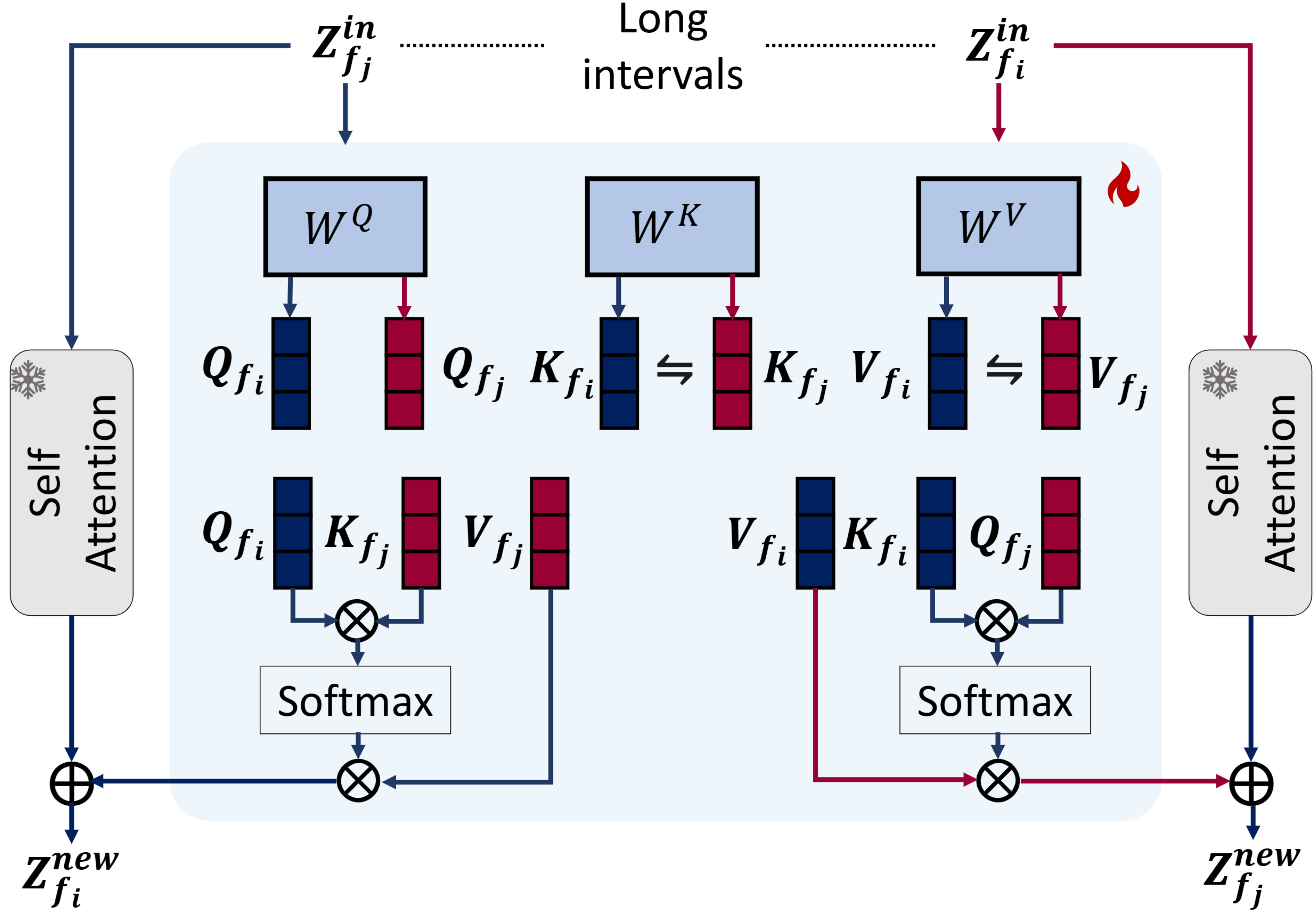}
    \vspace{-3mm}
    \caption{
    Bidirectional Attention. We inject bidirectional attention layers parallel to self-attention layers in diffusion models. 
    During training, we specifically optimize the bidirectional attention layer while keeping the other layers frozen.
    }
    \label{fig:method-B}
    \vspace{-3mm}
\end{figure}

\begin{figure}[tp]
    \centering
    \includegraphics[width=.9\linewidth]{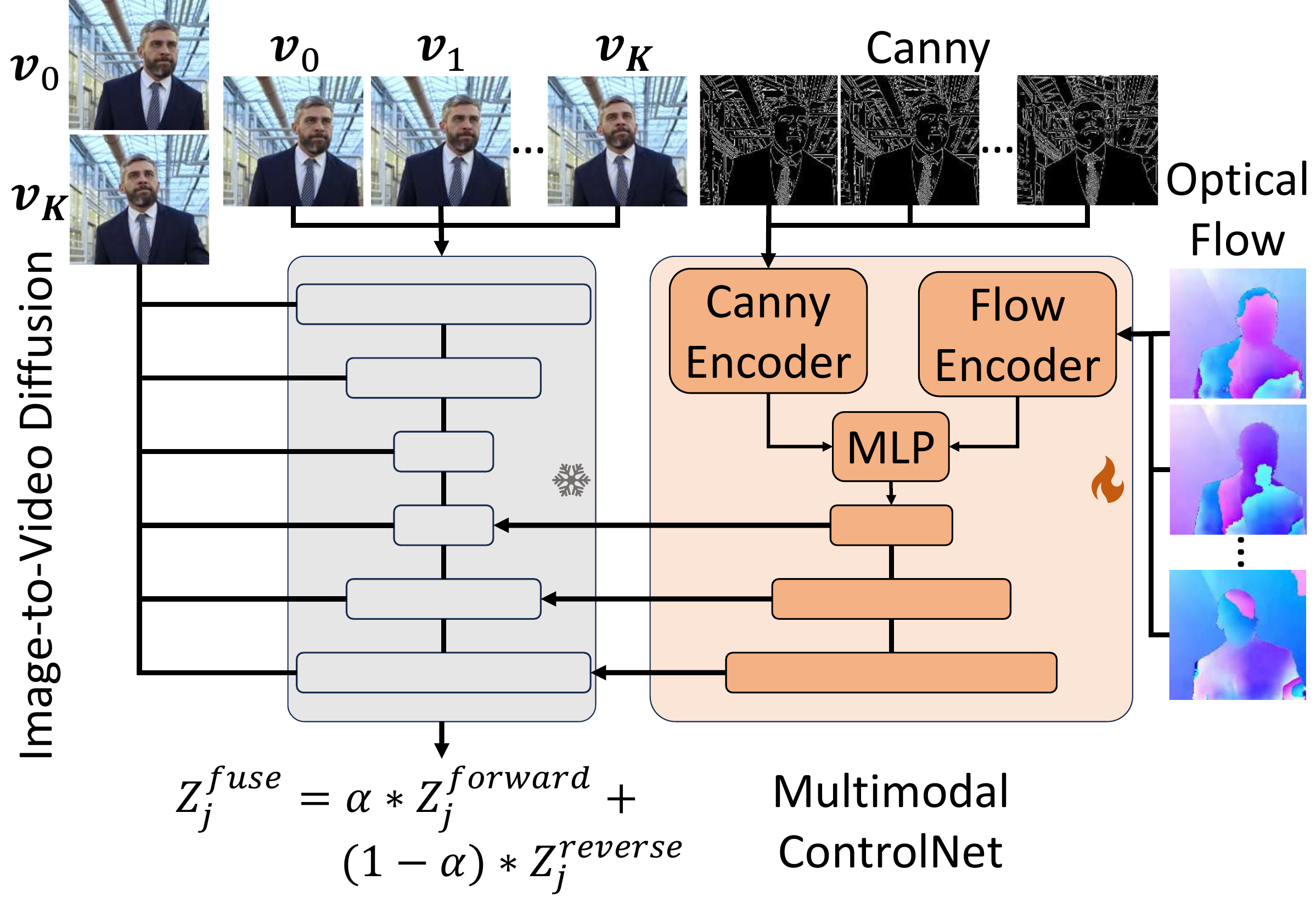}
    \vspace{-3mm}
    \caption{Multimodal ControlNet. Using Canny edges and optical flow to combine static structure and motion dynamics.
    % We input both Canny edge and optical flow to inject static structure and motion dynamics simultaneously.
    % We train the multimodal ControlNet for video diffusion models, which simultaneously receives optical flow and Canny edge inputs as supervision for motion and structure.
    }
    \vspace{-3mm}
    \label{fig:method-C}
\end{figure}

\vspace{-1mm}
\subsection{Anchor Editing via Pairwise Diffusion}

Editing an entire video sequence frame by frame is computationally expensive and prone to producing temporally inconsistent results~\cite{t2vedit-streamv2v,i2vedit-flowvid}. To address this, we adopt a two-stage strategy by first editing a sparse set of anchor frames that capture key semantic and structural variations throughout the video.

Following previous works~\cite{zeroshot-rerender,i2vedit-flowvid}, we uniformly sample anchor frames every $K$ frames, forming a set $\boldsymbol{F} = \{\boldsymbol{f}_i\}_{i=0}^M$ where $\boldsymbol{f}_i = \boldsymbol{v}_{i \cdot K}$. These anchors are expected to provide global coverage in both motion and appearance, serving as the backbone for subsequent propagation.

A simple approach is to edit each anchor frame independently using an off-the-shelf image editing model conditioned on the original prompt $\boldsymbol{y}^{inv}$ and the editing prompt $\boldsymbol{y}^{edit}$ and optional structural cues. This can be formulated as:
$\boldsymbol{f}_i' = \mathcal{E}(\boldsymbol{f}_i,  \boldsymbol{y}^{inv}, \boldsymbol{y}^{edit}, c)$,
where $\mathcal{E}$ is a text-to-image diffusion model and $c$ denotes additional control signals such as edges or depth. However, such frame-wise editing lacks inter-frame awareness and often results in visual inconsistency across anchors, as shown in Figure ~\ref{fig:ablation-pairwise}.

To produce coherent long-range edits, it is essential to move beyond independent frame processing and instead treat anchor frames as a set of mutually conditioned entities. Rather than optimizing each frame in isolation, the editing process should incorporate global context, allowing each frame to be influenced by others in the set. This motivates a new formulation that extends traditional diffusion editing to a joint process over multiple frames—a direction we refer to as \textbf{Pairwise Diffusion}.

{\flushleft\bf Bidirectional Attention.} While anchor frame sampling reduces the scale of the editing task, achieving consistency across the sampled frames remains non-trivial. Editing each frame individually fails to establish correlations between frames, resulting in misaligned structure or semantics. To explicitly model such cross-frame dependencies, we introduce a \textbf{Bidirectional Attention} mechanism into the diffusion backbone (Figure ~\ref{fig:method-B}).
This module allows each anchor frame to attend to another anchor frame during denoising, enabling mutual exchange of structural and semantic information. The attention is injected into each layer of the denoising U-Net and operates in parallel with the standard self-attention.

Formally, let $\boldsymbol{Z}^{\text{in}}_{\boldsymbol{f}_i}$ and $\boldsymbol{Z}^{\text{in}}_{\boldsymbol{f}_j}$ denote the latent features of two anchor frames. We first compute queries, keys, and values for both:
$
\boldsymbol{Q}_{\boldsymbol{f}_i}, \boldsymbol{K}_{\boldsymbol{f}_i}, \boldsymbol{V}_{\boldsymbol{f}_i}, \quad 
\boldsymbol{Q}_{\boldsymbol{f}_j}, \boldsymbol{K}_{\boldsymbol{f}_j}, \boldsymbol{V}_{\boldsymbol{f}_j},
$
using shared linear projections to maintain symmetry. Bidirectional attention is then computed in both directions:
\begin{align}
    \boldsymbol{Z}^{\text{out}}_{\boldsymbol{f}_i} &= \text{Softmax}\left( \frac{\boldsymbol{Q}_{\boldsymbol{f}_i} \boldsymbol{K}_{\boldsymbol{f}_j}^\top}{\sqrt{d}} \right) \boldsymbol{V}_{\boldsymbol{f}_j}, \\
    \boldsymbol{Z}^{\text{out}}_{\boldsymbol{f}_j} &= \text{Softmax}\left( \frac{\boldsymbol{Q}_{\boldsymbol{f}_j} \boldsymbol{K}_{\boldsymbol{f}_i}^\top}{\sqrt{d}} \right) \boldsymbol{V}_{\boldsymbol{f}_i}.
\end{align}

The output of the Bidirectional Attention is fused with the original self-attention result:
\begin{align}
    \boldsymbol{Z}^{\text{new}}_{\boldsymbol{f}_i} &= \text{SelfAttn}(\boldsymbol{Z}^{\text{in}}_{\boldsymbol{f}_i}) + \boldsymbol{Z}^{\text{out}}_{\boldsymbol{f}_i}, \\
    \boldsymbol{Z}^{\text{new}}_{\boldsymbol{f}_j} &= \text{SelfAttn}(\boldsymbol{Z}^{\text{in}}_{\boldsymbol{f}_j}) + \boldsymbol{Z}^{\text{out}}_{\boldsymbol{f}_j}.
\end{align}

This formulation ensures that both frames incorporate complementary signals from each other during generation, enabling alignment in both appearance and structure. As shown in prior works~\cite{liu2024towards, xiao2024fastcomposer, guan2025hybridbooth, imagefeature-pnp}, enhancing attention pathways with cross-instance context improves generation fidelity. Our design follows this principle to maintain coherent editing across anchor frames.

{\flushleft\bf Progressive Pairwise Diffusion with Feature Injection.} 
With Bidirectional Attention serving as the foundation for cross-frame information exchange, we now extend the editing process from isolated frame pairs to the full set of anchor frames. Inspired by ~\cite{gen-l-video,collaborative-video}, we adopt a progressive denoising framework that iteratively edits all neighboring anchor frame pairs. Specifically, for each adjacent pair $(\boldsymbol{f}_i, \boldsymbol{f}_{i+1})$, we perform joint denoising across $T$ steps. For frames that appear in multiple overlapping pairs (e.g., $\boldsymbol{f}_1$ in both $(\boldsymbol{f}_0, \boldsymbol{f}_1)$ and $(\boldsymbol{f}_1, \boldsymbol{f}_2)$), we average the intermediate latents at each denoising step $t$ which are called \textbf{cross-pair fusion}:
\begin{equation}
\boldsymbol{x}_t^{\boldsymbol{f}_1} = \frac{1}{2} \left( \mathbf{e}_t(\boldsymbol{f}_0, \boldsymbol{f}_1)[1] + \mathbf{e}_t(\boldsymbol{f}_1, \boldsymbol{f}_2)[0] \right),
\end{equation}
where $\mathbf{e}_t(\cdot, \cdot)$ denotes the denoising output at step $t$ for a given anchor frame pair.
This progressive averaging encourages smooth transitions and structural consistency across long sequences. 

To preserve spatial structure and visual identity during editing, we incorporate a Plug-and-Play (PnP) style feature injection mechanism~\cite{imagefeature-pnp}, which decouples structural preservation from semantic transformation. Anchor frames are first inverted into the latent space via pairwise diffusion guided by an inversion prompt $\boldsymbol{y}^{inv}$, capturing the original content of the video.
During the editing process, frames are generated based on the editing prompt $\boldsymbol{y}^{edit}$, while key and value features from the inversion trajectory are injected into the attention and conv layers of the editing stream. This dual-prompt guidance enables the model to apply semantic changes while retaining spatial layout and structural coherence from the original video.

While Bidirectional Attention facilitates interaction between frame pairs, we observe that its effect can be diminished when frame similarity is weak or noisy. To address this, we introduce a \textbf{multi-conditional guidance scheme} that enhances the influence of both semantic and structural conditions during generation. Inspired by composable diffusion~\cite{multicfg} and instruction-based editing~\cite{instructpix2pix}, we modify the denoising model $\mathbf{e}_\theta(\boldsymbol{x}_t, c_T, c_J)$ to support two conditions: a text prompt $c_T$ and a joint structural condition $c_J$. The final guided output is computed as:
\begin{equation}
\begin{aligned}
  \tilde{\mathbf{e}}_\theta(\boldsymbol{x}_t, c_T, c_J) =\ 
  &\mathbf{e}_\theta(\boldsymbol{x}_t, \varnothing, c_J)\ + \\
  &s_T \cdot \left( \mathbf{e}_\theta(\boldsymbol{x}_t, c_T, c_J) - \mathbf{e}_\theta(\boldsymbol{x}_t, \varnothing, c_J) \right)\ + \\
  &s_J \cdot \left( \mathbf{e}_\theta(\boldsymbol{x}_t, c_T, c_J) - \mathbf{e}_\theta(\boldsymbol{x}_t, c_T, \varnothing) \right),
\end{aligned}
\label{eq:joint guidance}
\end{equation}
where $s_T$ and $s_J$ are tunable guidance scales. This formulation amplifies the impact of the editing prompt and the joint structural condition, respectively. Combined with progressive pairwise denoising and structural feature injection, our guidance scheme enables globally consistent, structure-preserving edits across all anchor frames in long video sequences.

\begin{table}[t]
\centering
\resizebox{\columnwidth}{!}{%
\begin{tabular}{lccc}
\toprule
 & Short (\(< 5\text{s}\)) & Medium (\(5\text{s} \sim 30\text{s}\)) & Long (\(> 30\text{s}\)) \\
\midrule
\# Videos & 20 & 20 & 20 \\
\# Avg. Frames & 76 & 375 & 1500 \\
\# Prompts & 90 & 94 & 100 \\
\bottomrule
\end{tabular}
}
\caption{Statistics of our evaluation dataset.
% Number of videos and prompts across different durations.
}
%\vspace{-3mm}
\label{tab:data}
\vspace{-4mm}
\end{table}

\vspace{-1mm}
\subsection{Multimodal-Guided Interpolation}

After achieving globally consistent edits on the anchor frames, the next step is to synthesize the intermediate frames between each anchor pair while preserving both temporal dynamics and spatial fidelity. Given two adjacent edited anchors $\boldsymbol{f}'_{i}$ and $\boldsymbol{f}'_{i+1}$ (corresponding to $\boldsymbol{v}'_{i \cdot K}$ and $\boldsymbol{v}'_{(i+1) \cdot K}$), our goal is to reconstruct the intermediate sequence as follows:
\begin{equation} \label{eq:interpolate}
\{\boldsymbol{v}'_{j}\}_{j=i \cdot K + 1}^{(i+1) \cdot K - 1} = \mathcal{I}(\boldsymbol{v}'_{i \cdot K}, \boldsymbol{v}'_{(i+1) \cdot K} \mid \{\boldsymbol{v}_{j}\}_{j=i \cdot K}^{(i+1) \cdot K}),
\end{equation}
where $\mathcal{I}(\cdot)$ denotes the interpolation module, and $\{\boldsymbol{v}_{j}\}$ is the corresponding segment from the original video.

Previous methods such as Rerender~\cite{zeroshot-rerender}, which use Ebsynth~\cite{interpolate-ebsynth} guided by optical flow to warp frames between anchors, often struggle with fast or complex motion, leading to visible artifacts. Other interpolation-based approaches~\cite{interpolate-diffmorpher, interpolate-dreammover} similarly suffer from unrealistic transitions and loss of spatial coherence. These issues highlight the challenge of maintaining both temporal smoothness and structural consistency in long-range interpolation.

To address this, we adopt a diffusion-based image-to-video generation model as the core of our interpolation module $\mathcal{I}$. Recent works such as Stable Video Diffusion~\cite{i2v-svd} demonstrate the capacity of such models to synthesize temporally coherent and visually realistic video sequences. However, applying this model directly for interpolation requires addressing two key constraints: (1) both the start and end anchor frames should guide generation, not just one; (2) additional guidance is needed to preserve motion dynamics and spatial layout from the original content.

{\flushleft\bf Bidirectional Temporal Frame Fusion.} 
Standard image-to-video generation proceeds in a forward manner, synthesizing frames sequentially from a starting image, e.g., $\boldsymbol{v}_{i \cdot K} \rightarrow \boldsymbol{v}_{(i+1) \cdot K}$. This often leads to cumulative drift, and the final frame may deviate from the desired target anchor. To alleviate this, we incorporate a symmetric reverse generation process, where the model also synthesizes frames in reverse, i.e., $\boldsymbol{v}_{i \cdot K} \leftarrow \boldsymbol{v}_{(i+1) \cdot K}$. Both directions are initialized from shared noise and conditioned on their respective starting anchors.

To ensure the outputs from both directions remain consistent with their boundary frames, we linearly blend their latent representations at each timestep. Specifically, let $\boldsymbol{Z}^{\text{forward}}_j$ and $\boldsymbol{Z}^{\text{reverse}}_j$ denote the latent representations of the $j$-th frame from the forward and reverse passes, respectively. The final fused latent is computed as:
\begin{equation}
    \boldsymbol{Z}^{\text{fuse}}_j = \alpha \boldsymbol{Z}^{\text{forward}}_j + (1-\alpha) \boldsymbol{Z}^{\text{reverse}}_j,
\end{equation}
where $\alpha$ is a linearly decreasing weight from $1$ to $0$, ensuring a smooth transition between the two trajectories.

\begin{figure*}[t]
    \centering
    \includegraphics[width=.96\linewidth]{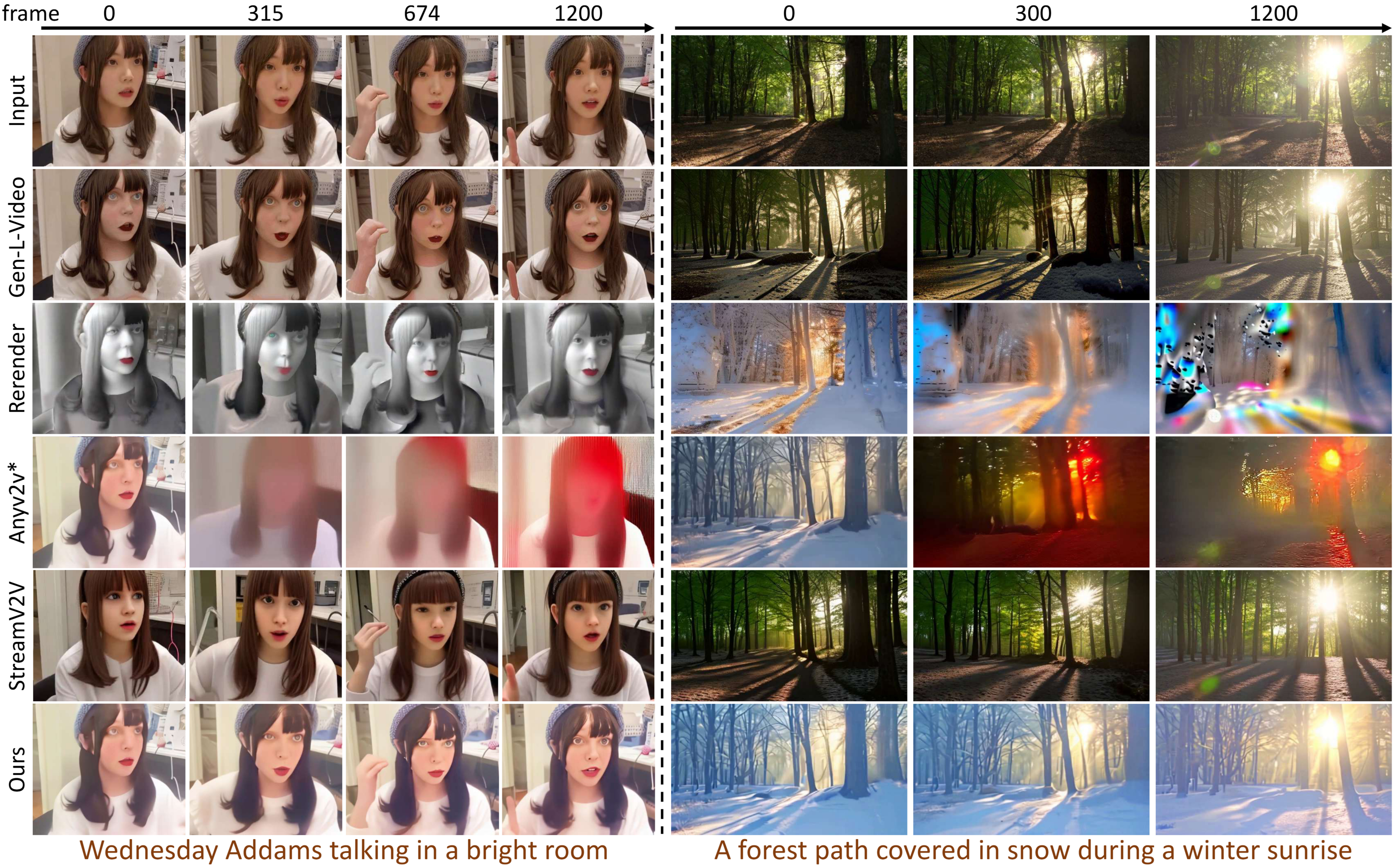}
    \vspace{-4mm}
    \caption{Qualitative results. Our method demonstrates excellent performance in temporal continuity and structural consistency. 
    }
    \label{fig:Qualitative}
    %\vspace{-0.3cm}
\end{figure*}

{\flushleft\bf Multimodal Conditioning.}
While bidirectional fusion aligns temporal endpoints, it alone cannot preserve detailed motion patterns or spatial structure specific to the original video. To enhance control, we adopt a multimodal ControlNet~\cite{controlnet} as an auxiliary guidance branch in both generation streams.

We utilize two types of visual cues: Canny edge maps, which represent static structure, and optical flow fields, which encode motion (Figure ~\ref{fig:method-C}). Each modality is processed by an independent encoder. The resulting features are concatenated and passed through a three-layer MLP, following the fusion design used in~\cite{xiao2024fastcomposer, guan2025hybridbooth}. The merged features are then injected into a frozen diffusion U-Net, serving as soft control signals during the denoising steps.

This multimodal conditioning ensures that the synthesized frames adhere to both the edited semantics and the original video’s structural and motion constraints, enabling temporally smooth and spatially consistent interpolation across segments.

\begin{table*}[t]
\centering
\resizebox{.86\linewidth}{!}{%
\begin{tabular}{lcccccccc}
\toprule
& \multicolumn{2}{c}{\textbf{Long-Term Metric}} & \multicolumn{2}{c}{\textbf{Frame Continuity}} & \multicolumn{2}{c}{\textbf{Structure Consistency}} & \textbf{Frame Quality} & \textbf{Text Alignment} \\
\cmidrule(lr){2-3}\cmidrule(lr){4-5}\cmidrule(lr){6-7}\cmidrule(lr){8-8}\cmidrule(lr){9-9}
& I-I CLIP Sim$^\star$ $^\uparrow$ & I-I CLIP Sim$^\dagger$ $^\uparrow$ & I-I CLIP Sim. $^\uparrow$ & I-I DINO Sim. $^\uparrow$ & Warp Error $^\downarrow$ & Canny Error $^\downarrow$ & Entropy $^\uparrow$ & I-T CLIP Sim. $^\uparrow$ \\
\midrule
Rerender \cite{zeroshot-rerender} & 95.57 & 91.22 & 99.36 & 99.43 & 9.34 & 11.21 & 6.68 & 18.97 \\
Gen-L-Video \cite{gen-l-video} & 96.74· & 92.25 & 99.19 & 99.29 & 8.16 & 10.82 & 7.16 & 18.37 \\
Anyv2v* \cite{i2vedit-anyv2v} & 94.54 & 88.21 & 99.02 & 98.23 & 5.19 & 14.23 & 4.35 & 17.22 \\
StreamV2V \cite{t2vedit-streamv2v} & 94.23 & 90.54 & 98.81 & 99.08 & 7.41 & 10.81 & 7.11 & 18.59 \\
\midrule
\textbf{Ours} & \textbf{97.84} & \textbf{94.11} & \textbf{99.64} & \textbf{99.67} & \textbf{3.78} & \textbf{10.36} & \textbf{7.42} & \textbf{19.16} \\
\bottomrule
\end{tabular}
}
% \vspace{-0.2cm}
\caption{
Quantitative comparison. StreamingV2V edits long videos frame-by-frame, while others edit segment-by-segment.
% The best results are \textbf{highlighted} in \textbf{bald}.
}
\vspace{-4mm}
\label{tab:quantitative}
\end{table*}

\begin{table}[t]
\centering
\resizebox{\columnwidth}{!}{
\begin{tabular}{@{}l@{}c@{}c@{}c@{}c@{}}
\toprule
& Frame Continuity $^\uparrow$ & Structure Consistency $^\uparrow$ & Frame Quality $^\uparrow$ & Text Alignment $^\uparrow$ \\
\midrule
Rerender \cite{zeroshot-rerender}       & 1.44 & 0.72 & 1.57 & 1.76 \\
Gen-L-Video \cite{gen-l-video}          & 1.19 & 1.38 & 1.05 & 0.57 \\
Anyv2v* \cite{i2vedit-anyv2v}           & 0.65 & 0.29 & 0.23 & 0.18 \\
StreamV2V \cite{t2vedit-streamv2v}      & 0.27 & 0.73 & 2.03 & 2.56 \\
\textbf{Ours}                           & \textbf{4.45} & \textbf{4.88} & \textbf{3.12} & \textbf{2.93} \\
\bottomrule
\end{tabular}
}
\caption{Results of user studies.}
\vspace{-1cm}
\label{tab:quantitative-userstudy}
\end{table}

\vspace{-1mm}
\section{Experiments}
%\vspace{-0.1cm}
{\flushleft\bf Implementation Details.} In the anchor frame editing stage, we use \texttt{Stable Diffusion v1.5}~\cite{t2i-high} as the base image diffusion model and employ a Plug-and-Play (PnP) feature injection mechanism to preserve structural information during semantic editing. Specifically, key and value features extracted from the inversion stream are injected into the attention and convolutional layers of the editing stream, with injection ratios set to $0.44$ and $0.65$, respectively. We fine-tune the Bidirectional Attention module using Low-Rank Adaptation (LoRA)~\cite{lora} with rank and alpha set to $64$. Training samples are obtained by randomly selecting two frames from videos in the Panda-70M~\cite{pandas70m} dataset.

For interpolating intermediate frames, we adopt the \texttt{Stable Video Diffusion img2vid-xt} (SVD)~\cite{i2v-svd} model, augmented with a multimodal ControlNet similar to the architecture proposed in~\cite{multicontrolnet-cocktail}. Following SVD’s design, we replace the 2D convolution and spatial attention layers in ControlNet~\cite{controlnet} with 3D convolution and spatiotemporal attention to better capture video dynamics. Canny edges and optical flow are used as multimodal conditions. Each modality is encoded with a dedicated ResNet~\cite{he2016deep} encoder consisting of 5 convolutional layers, with input channels of 1 for edge maps and 2 for flow fields. The outputs are fused via a 5-layer MLP and injected into the diffusion backbone.

For joint guidance, we employ the multi-conditional guidance scheme described in Eq.~\ref{eq:joint guidance}, setting the text guidance scale $s_T = 6.0$ and the joint condition scale $s_J = 0.8$.
Anchor frames are selected every 24 frames to balance editing efficiency and interpolation quality, following the recommended temporal spacing in SVD. During training, we extract 25-frame clips from Panda-70M, and center-crop and resize all frames to a resolution of $576 \times 320$.
We use the AdamW optimizer~\cite{adam} with a fixed learning rate of $0.0001$ and $(\beta_1, \beta_2) = (0.9, 0.999)$ with 500-step warm-up. The Bidirectional Attention is trained for $90,000$ steps, and the Multimodal ControlNet for $70,000$ steps.

Our validation dataset covers short, medium, and long video sequences across a diverse set of scenarios including urban scenes, nature, humans, animals, and dynamic objects. All videos are standardized to $576 \times 320$ resolution and 24 FPS. A detailed breakdown of the dataset composition is provided in Tab.~\ref{tab:data}.

{\flushleft\bf Evaluation Metrics.}
We evaluate editing performance across several dimensions to comprehensively assess quality, consistency, and semantic alignment.

\textbf{(1) Long-Term Metric.}  
To quantify temporal consistency over extended durations, we introduce two CLIP-based metrics tailored for long video evaluation:  
(i) \textit{I-I CLIP Sim$^\star$} measures the average CLIP similarity between frames that are 24 frames apart, capturing stability over time.  
(ii) \textit{I-I CLIP Sim$^\dagger$} computes the similarity between every 24th frame and the first frame, highlighting overall coherence and identity preservation across the sequence.

\textbf{(2) Frame Continuity.}  
We evaluate short-term temporal smoothness by computing cosine similarity between adjacent frames using CLIP embeddings, following prior work~\cite{t2vedit-streamv2v, zeroshot-pix2video, oneshot-tuneavideo}. We denote this as \textit{I-I CLIP Sim.} We also report \textit{I-I DINO Sim.}, based on DINO~\cite{dino} features, to better reflect structural-level continuity.

\textbf{(3) Structural Consistency.}  
To evaluate the preservation of spatial structure, we compute the \textit{Warp Error}~\cite{warpErr} and the \textit{Canny Error}, which measure pixel misalignment and edge-level differences, respectively. These metrics follow the protocols of~\cite{t2vedit-streamv2v, vidtome} and are calculated between the original and edited frames.

\textbf{(4) Frame Quality.}  
To assess perceptual quality, especially sharpness and detail, we compute information entropy for each frame. Higher entropy indicates richer textures and less blur.

\textbf{(5) Text-Image Alignment.}  
To assess how well the generated frames reflect the editing prompt, we compute the cosine similarity between the CLIP embeddings of each frame and the corresponding textual description, denoted as \textit{I-T CLIP Sim.}.

Additional implementation and normalization details for all metrics are included in the Supplementary Material.

% \vspace{-0.2cm}
{\flushleft\bf User Study.}  
We further conduct a user study to evaluate visual quality from a human perceptual standpoint. A group of 60 graduate participants was asked to rank outputs from our method and competing baselines on 100 randomly selected edited videos out of a pool of 284. Final scores are computed based on normalized ranking preferences across users. More details are available in the Supplementary Material.

%\vspace{-0.35cm}
{\flushleft\bf Compared Methods.}
We compare our method against several recent baselines, including Rerender~\cite{zeroshot-rerender}, Gen-L-Video~\cite{gen-l-video}, Anyv2v~\cite{i2vedit-anyv2v}, and StreamV2V~\cite{t2vedit-streamv2v}, covering both frame-by-frame and segment-by-segment editing paradigms.
StreamV2V performs editing in a streaming fashion, updating the video frame-by-frame.
Rerender edits sparse keyframes and interpolates intermediate frames using Ebsynth.
Gen-L-Video divides the video into overlapping segments and applies local editing with latent-space blending.
Anyv2v generates the output conditioned on the first edited frame; we adapt it (Anyv2v*) for segment-wise editing by conditioning each segment on the final frame of the previous segment.
All baseline methods are evaluated using their publicly available implementations and default hyperparameters.

\vspace{-1mm}
\subsection{Comparison with State-of-the-Arts}

{\flushleft\bf Quantitative Evaluation.}
Quantitative results and user studies are presented in Tables~\ref{tab:quantitative} and~\ref{tab:quantitative-userstudy}. 
\textbf{Gen-L-Video} maintains relatively good global consistency through segment-wise blending. However, its overall structural consistency and frame continuity remain limited, mirroring the inherent limitations of the short-video editing methods it builds upon.
\textbf{Rerender} performs well in general but shows weaker structural consistency, particularly in terms of warp error and long-term alignment, due to inconsistencies in anchor frame editing and the lack of global context modeling.
\textbf{Anyv2v*} underperforms on all metrics due to cumulative errors that lead to progressive artifacts and structural distortions.
\textbf{StreamV2V} benefits from a streaming editing pipeline and cached historical features, but it cannot ensure strong global structure. It often produces subtle flickering or warping, especially in high-frequency details, limiting its temporal smoothness.
\textbf{Our method} outperforms all baselines across most quantitative metrics, particularly in long-term consistency, frame continuity, and structural coherence. It also receives the highest ratings in user studies, demonstrating superior perceptual quality and overall reliability.

%\vspace{-0.3cm}
{\flushleft\bf Qualitative Evaluation.}
% 先做这个
We show qualitative comparison results in Fig.~\ref{fig:Qualitative}.
\textbf{Gen-L-Video} leverage latent blending across overlapping segments, which helps preserve overall consistency. However, it still inherits limitations such as temporal inconsistency and occasional visual stutters due to the limitations of short video methods. 
% These issues are especially evident in challenging cases, where identity preservation and motion smoothness are critical.
\textbf{Rerender} retains motion well but suffers from noticeable inconsistency in details, such as clothes color, facial structure, and snow. Its keyframe-based pipeline with Ebsynth interpolation leads to blurry transitions and misalignments when large changes occur.
\textbf{Anyv2v*} exhibits noticeable degradation in long videos due to accumulated errors across segments. Its frame-to-segment conditioning leads to progressive blurring, structural collapse, and identity drift, especially in later portions of the sequence.
\textbf{StreamV2V} provides relatively stable results but still shows flickering and spatial inconsistencies, particularly in the face and hair regions. 
% These artifacts stem from the limited capacity of its cached-history mechanism to model long-term dependencies.
In contrast, \textbf{our method} produces temporally stable and structurally coherent results across all frames. It preserves identity, motion, and appearance details effectively, even in long sequences. For visual results, we recommend watching the supplementary video.

\vspace{-1mm}
\subsection{Ablation Studies}
%设计ablation

\begin{figure}[t]
    \centering
    \includegraphics[width=.9\linewidth]{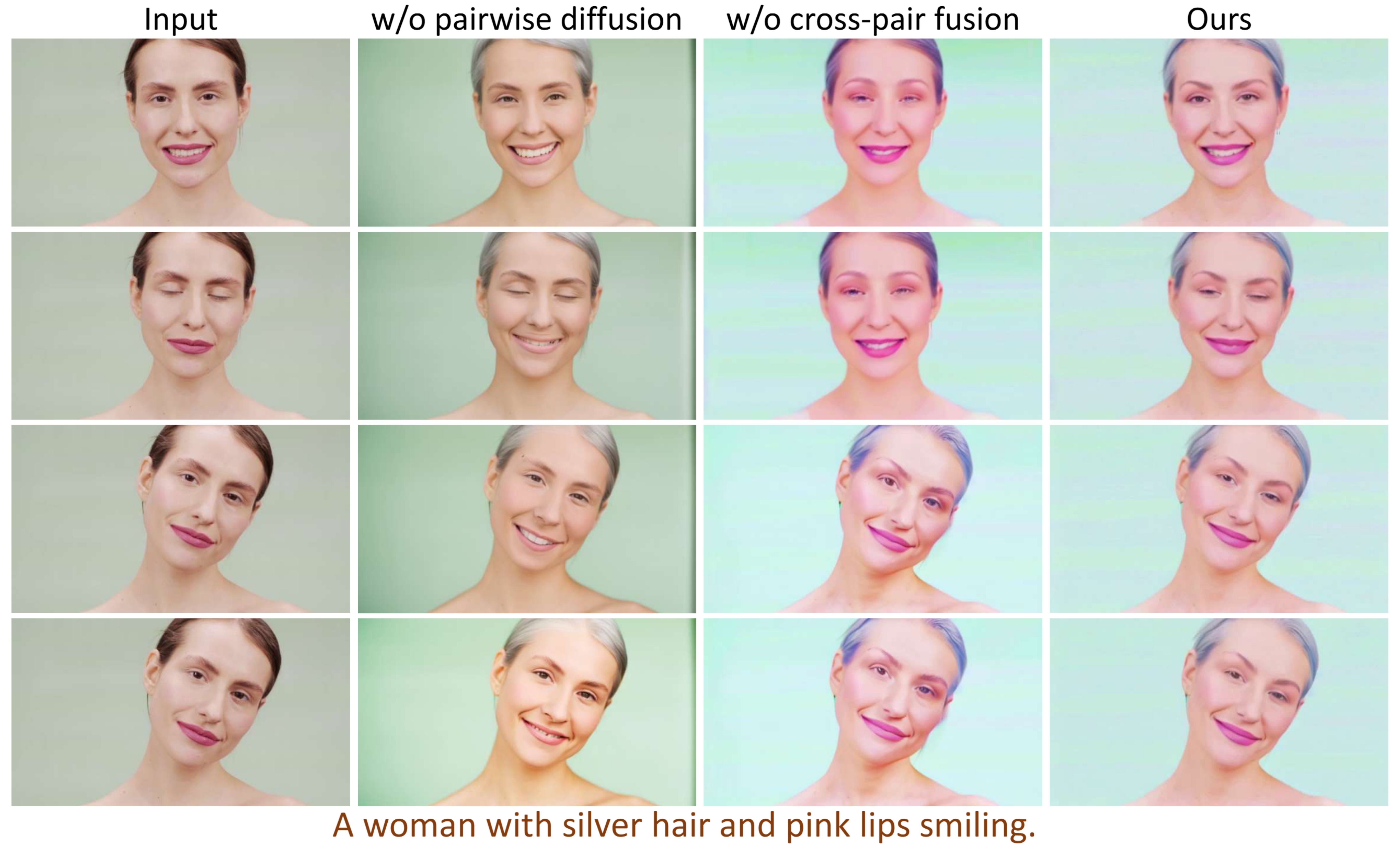}
    \vspace{-3mm}
    \caption{Analysing on Pairwise Diffusion.}
    \label{fig:ablation-pairwise}
    \vspace{-2mm}
\end{figure}

%\vspace{-0.1cm}
{\flushleft\bf Analyses on Pairwise Diffusion.}
We perform ablations to analyze the contributions of the two key components in our pairwise diffusion strategy: pairwise denoising and cross-pair fusion.
As illustrated in Fig.~\ref{fig:ablation-pairwise}, removing pairwise diffusion leads to frame-wise independent editing, where each anchor frame is modified in isolation. This results in clear inconsistencies across frames, especially in facial structure and appearance attributes such as lip color or hair shading.
When pairwise denoising is preserved but fusion across pairs is removed, the model edits each adjacent pair of frames (e.g., frame 1+2, 3+4) jointly. While local consistency within each pair is maintained, the absence of information flow across different pairs leads to a lack of global coherence. In other words, consistency is limited to short temporal spans, and transitions between different pairs can be visually abrupt.
In contrast, our full method employs progressive pairwise fusion, which averages shared latent features between overlapping pairs at each denoising step, enabling consistent editing propagation across the entire sequence.

\begin{figure}[t]
    \centering
    \includegraphics[width=0.8\linewidth]{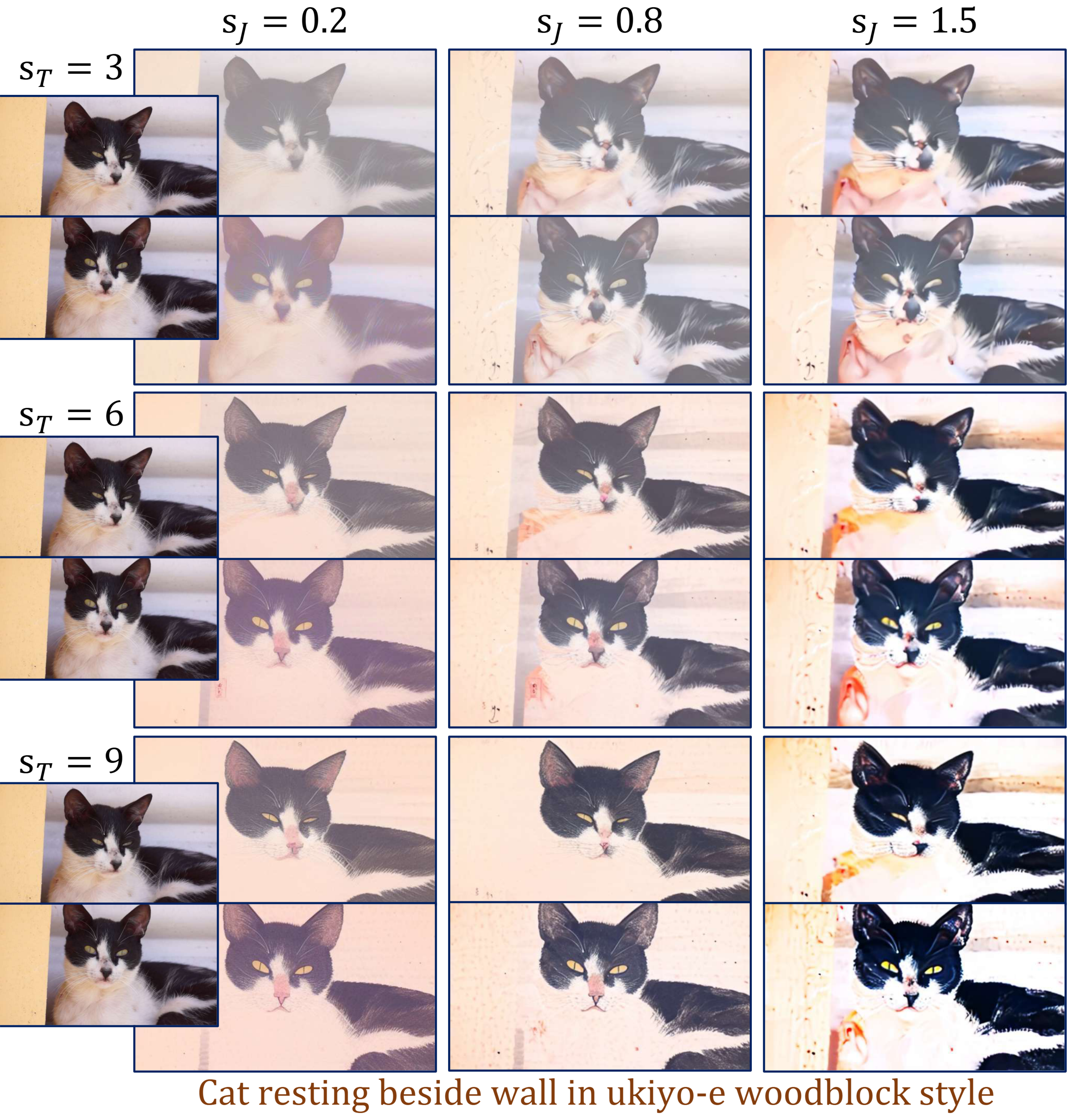}
    \vspace{-3mm}
    \caption{The impact of text guidance scale $s_T$ and joint guidance scale $s_J$. 
}
    \label{fig:ablation-scale_param}
    \vspace{-2mm}
\end{figure}

%\vspace{-0.2cm}
{\flushleft\bf Impact of Multi-conditional Guidance Scale.}
We study the influence of \(s_T\) (text guidance scale) and \(s_J\) (joint structural guidance) in our multi-conditional diffusion framework. As shown in Fig.~\ref{fig:ablation-scale_param}, increasing \(s_T\) enhances semantic transformation strength, producing more pronounced edits. In contrast, higher \(s_J\) improves global coherence across anchor frames by reinforcing structural alignment. However, overly strong guidance can cause unrealistic textures, oversaturation, and a loss of visual fidelity. In practice, we find that setting \(s_T = 6.0\) and \(s_J = 0.8\) strikes a good balance between edit fidelity and structural consistency.

\begin{figure}[t]
    \centering
    \includegraphics[width=.98\linewidth]{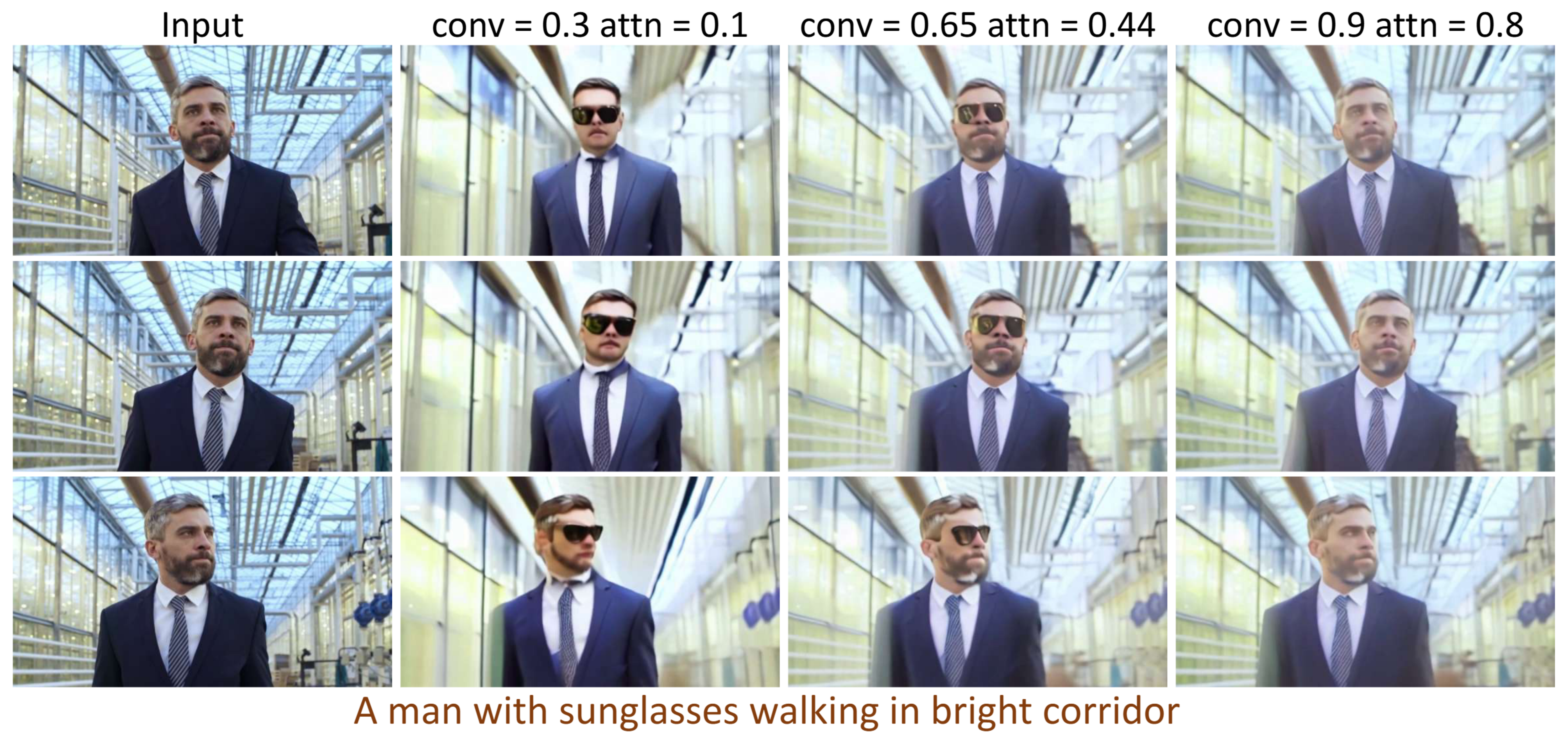}
    \vspace{-4mm}
    \caption{
    \textbf{Ablation on injection strength.}  
% We compare different feature injection settings for convolutional and attention layers.  
% Weak injection leads to structural distortion and inconsistency, while strong injection suppresses editing.  
Our default setting (middle) achieves the best trade-off.
}
    \label{fig:ablation-injection}
%    \vspace{-0.2cm}
\end{figure}

%\vspace{-0.2cm}
{\flushleft\bf Impact of Injection Ratios.}
To investigate the impact of Plug-and-Play feature injection on editing performance, we perform an ablation study by varying the injection ratios for convolutional and attention layers.
As shown in Fig.~\ref{fig:ablation-injection}, when the injection strength is too low (e.g., \texttt{conv} = 0.3, \texttt{attn} = 0.1), the model fails to preserve the structural identity of the subject, leading to shape distortions and temporal inconsistency. Conversely, when the injection is overly strong (e.g., \texttt{conv} = 0.9, \texttt{attn} = 0.8), the editing prompt has minimal influence, resulting in outputs that resemble the original input. 
Our default setting (\texttt{conv} = 0.65, \texttt{attn} = 0.44) provides a good trade-off, enabling effective semantic editing while maintaining structural coherence across frames. These results highlight the importance of balancing semantic transformation with structure retention in plug-and-play video editing pipelines.

\begin{table}[t]
\centering
\resizebox{.98\columnwidth}{!}{%
\begin{tabular}{c|cccc}
\toprule
\# Frame Interval & I-I CLIP Sim$^\star$$^\uparrow$ & I-I CLIP Sim. $^\uparrow$ & Warp Error $^\downarrow$ & Entropy $^\uparrow$ \\      
\midrule
48  & 97.11 & 99.31 & 4.28 & 7.02 \\
12  & 97.42 & 99.52 & 4.06 & 7.28 \\
6   & 97.23 & 99.48 & 4.55 & 7.21 \\
Random (6--24) & 97.04 & 99.55 & 4.14 & 7.31\\
w/o Interpolation & 96.81 & 99.39 & 5.86 & 7.18\\
\midrule
24  & 97.84 & 99.64 & 3.78 & 7.42 \\
\bottomrule
\end{tabular}
}
 % \vspace{-0.2cm}
\caption{Comparison of anchor frame selection strategies.}
% \vspace{-0.3cm}
\label{tab:anchor-frame-selection}
\vspace{-0.6cm}
\end{table}

%\vspace{-0.2cm}
{\flushleft\bf Anchor Frame Selection.}
We conduct an ablation study on anchor frame selection strategies, as shown in Tab.~\ref{tab:anchor-frame-selection}. Since SVD accepts up to 24 input frames, using intervals larger than 24 (e.g., 48-frame spacing in Line 2) results in suboptimal performance due to insufficient supervision during interpolation. Conversely, overly small intervals (Lines 3–4) increase the number of anchor frames, amplifying minor inconsistencies and leading to noticeable flickering in the output. Unevenly spaced anchors (Line 5) also exhibit slightly worse performance, likely due to unbalanced interpolation spans.
Although learning-based keyframe selection methods~\cite{keyframe-1,keyframe-2,keyframe-3,keyframe-4} could potentially improve coverage and edit propagation, they introduce additional computation and complexity. We leave the exploration of adaptive anchor selection as future work.

\begin{figure}[t]
    \centering
    \includegraphics[width=.98\linewidth]{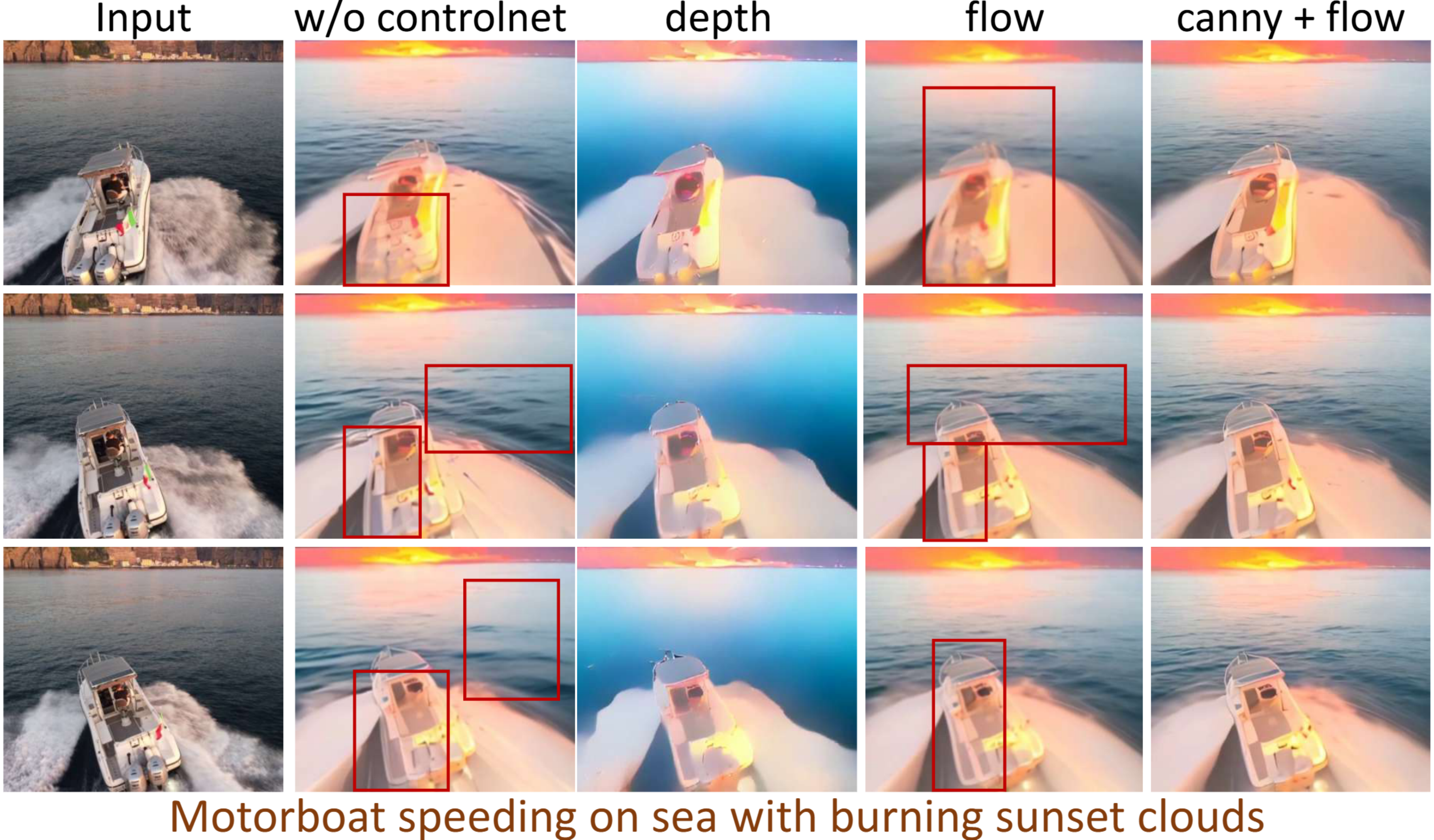}
    \vspace{-3mm}
    \caption{The performance of different ControlNets. Obvious inconsistencies are marked in red boxes.}
    \label{fig:ablation-multicontrol}
    \vspace{-2mm}
\end{figure}

%\vspace{-0.2cm}
{\flushleft\bf Multimodal Conditioning Interpolation.}
We evaluate the effect of different control signals in the interpolation stage, as shown in Fig.~\ref{fig:ablation-multicontrol}.
Without ControlNet, the image-to-video diffusion model lacks explicit motion guidance, which is supported by line 6 of Tab.~\ref{tab:anchor-frame-selection}. 
% While existing methods \cite{i2vedit-flowvid,i2vedit-ccedit,zeroshot-text2video} use depth as a control signal to maintain motion continuity, our results show that optical flow better preserves the motion of the subject.
While existing methods \cite{i2vedit-flowvid,i2vedit-ccedit,zeroshot-text2video} use depth as a control signal to maintain motion continuity, this often results in global motion blur and degraded structure preservation.
When using only optical flow, motion supervision is introduced, and subject dynamics are better preserved. However, in cases where foreground and background exhibit inconsistent motion or the scene contains complex structures, the flow-based control can produce artifacts such as jitter or regional blurring.
In contrast, our multimodal ControlNet combines flow with Canny edge information, enabling both motion and structural constraints. This fusion provides clearer spatial guidance and leads to temporally smooth and structurally consistent video interpolation results.

% {\flushleft\bf Impact of Inaccurate Optical Flow.} % 有没有都行，前面5个我觉得够了

\begin{figure}[t]
\centering
\includegraphics[width=.98\linewidth]{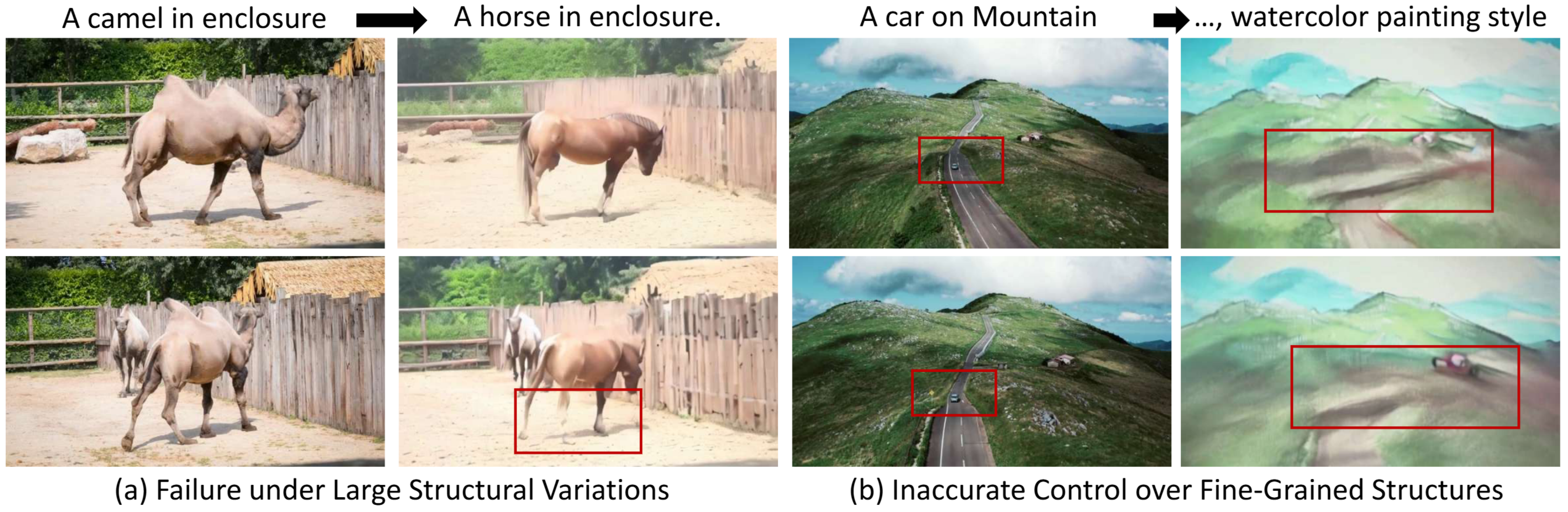}
\vspace{-3mm}
\caption{
\textbf{Limitations of our method.}
% 下面这段可以删掉
% \textbf{(a)} Failure under large structural variations. When the editing prompt causes drastic alterations in object shape, the model struggles to maintain temporal consistency across frames.  
% \textbf{(b)} Limited control over fine-grained structures. Our method may fail to accurately modify or preserve small-scale objects.
}
\label{fig:limitation}
\vspace{-0.5cm}
\end{figure}

\vspace{-1mm}
\section{Limitations}
% 看一遍结果图再决定limitation写什么
While our framework demonstrates strong performance on long video editing, it still faces several challenges. As shown in Fig.~\ref{fig:limitation}(a), when the editing involves drastic structural transformations—such as changing an object’s overall shape or category (e.g., camel to horse)—our method may fail to maintain temporal consistency. Even without rapid motion, significant semantic shifts can cause adjacent frames to diverge in structure or texture, resulting in visible flickering or identity drift. Additionally, as shown in Fig.~\ref{fig:limitation}(b), the model has difficulty controlling fine-grained structures. Editing small regions like facial expressions or tiny objects can be imprecise due to limited spatial resolution and weak feature localization in the current architecture. Nevertheless, our approach stands to benefit from future advances in image editing and video generation frameworks, particularly those enhancing structural reasoning and high-frequency detail preservation.

\vspace{-1mm}
\section{Conclusion}

% In this work, we present a novel two-stage framework for long video editing that achieves both semantic controllability and temporal consistency. We first perform anchor frame editing through a progressive pairwise diffusion strategy, enhanced by Plug-and-Play feature injection and Bidirectional Attention to enforce global coherence. The edited keyframes are then used to guide intermediate frame interpolation via a multimodal image-to-video generation module, leveraging both structural and motion cues through a ControlNet-based architecture.
% Extensive experiments demonstrate that our method outperforms existing approaches across various metrics, including temporal stability, structure preservation, and text alignment. Our design enables scalable and high-fidelity editing of minute-long videos, offering a practical solution to long video generation under user-defined semantics.
% We hope this work inspires future research on scalable, controllable video editing frameworks and encourages broader exploration of latent-space modeling across long temporal sequences.
We propose a two-stage framework for long video editing that combines globally consistent anchor frame editing with multimodal-guided interpolation. Our method integrates Plug-and-Play feature injection, Bidirectional Attention, and multi-conditional guidance to ensure semantic alignment and structural fidelity. Experimental results demonstrate clear advantages over prior work in long-range consistency and visual quality. This work provides a scalable solution for high-quality, controllable long video editing.

%%
%% The next two lines define the bibliography style to be used, and
%% the bibliography file.
% \newpage
%\vspace{-0.3cm}
\begin{acks}
This work was supported by NSFC (62322113, 62376156), Shanghai Municipal Science and Technology Major Project (2021SHZDZX0102), the Fundamental Research Funds for the Central Universities, and Ant Group.
\end{acks}

\bibliographystyle{ACM-Reference-Format}
\balance 
\bibliography{sample-base}

%%
%% If your work has an appendix, this is the place to put it.
% \appendix

% \section{Research Methods}

% \subsection{Part One}

% Lorem ipsum dolor sit amet, consectetur adipiscing elit. Morbi
% malesuada, quam in pulvinar varius, metus nunc fermentum urna, id
% sollicitudin purus odio sit amet enim. Aliquam ullamcorper eu ipsum
% vel mollis. Curabitur quis dictum nisl. Phasellus vel semper risus, et
% lacinia dolor. Integer ultricies commodo sem nec semper.

% \subsection{Part Two}

% Etiam commodo feugiat nisl pulvinar pellentesque. Etiam auctor sodales
% ligula, non varius nibh pulvinar semper. Suspendisse nec lectus non
% ipsum convallis congue hendrerit vitae sapien. Donec at laoreet
% eros. Vivamus non purus placerat, scelerisque diam eu, cursus
% ante. Etiam aliquam tortor auctor efficitur mattis.

% \section{Online Resources}

% Nam id fermentum dui. Suspendisse sagittis tortor a nulla mollis, in
% pulvinar ex pretium. Sed interdum orci quis metus euismod, et sagittis
% enim maximus. Vestibulum gravida massa ut felis suscipit
% congue. Quisque mattis elit a risus ultrices commodo venenatis eget
% dui. Etiam sagittis eleifend elementum.

% Nam interdum magna at lectus dignissim, ac dignissim lorem
% rhoncus. Maecenas eu arcu ac neque placerat aliquam. Nunc pulvinar
% massa et mattis lacinia.

\end{document}